\documentclass[runningheads]{llncs}
\usepackage{graphicx}

\usepackage[dvipsnames]{xcolor}
\usepackage{tikz}
\usepackage{pgfplots}
\usepackage{pgfplotstable}
\pgfplotsset{compat=1.17}
\usetikzlibrary{pgfplots.groupplots}
\usetikzlibrary{calc,arrows}
\usepackage{comment}
\usepackage{amsmath,amssymb} 
\usepackage{color}
\usepackage{colortbl}
\usepackage{wrapfig}
\usepackage{graphicx}
\usepackage{float}
\usepackage{subfig}
\usepackage[hyphens]{url}
\usepackage[export]{adjustbox}

\pgfplotsset{ every non boxed x axis/.append style={x axis line style=-},
	every non boxed y axis/.append style={y axis line style=-}}

\usepackage[pagebackref,breaklinks,colorlinks,citecolor=eccvblue]{hyperref}
\usepackage{booktabs} 
\usepackage{tabularx}
\usepackage{multirow}
\usepackage{makecell}

\usepackage{xspace}
\usepackage{longtable}
\usepackage{algpseudocode}
\usepackage{listings}
\usepackage[ruled]{algorithm2e}
\SetAlCapFnt{\scriptsize}

\usepackage{eccv}
\usepackage{eccvabbrv}

\usepackage{orcidlink}
\usepackage{utfsym}
\usepackage{soul}
\usepackage{wrapfig}
\usepackage{marvosym}

\begin{document}
	\pagestyle{headings}
	\mainmatter
	\def\ECCVSubNumber{7436}  
	
	\title{Diffusion for Natural Image Matting}
	

        \author{Yihan Hu\inst{1,2,5}\orcidlink{0009-0004-1804-0518} \and
        Yiheng Lin\inst{1,2}\orcidlink{0009-0008-5399-59684} \and
        Wei Wang\inst{1,2}\orcidlink{0000-0002-5477-1017} \and 
        Yao Zhao\inst{1,2,3}\orcidlink{0000-0002-8581-9554}\\
        Yunchao Wei\inst{1,2,3}\textsuperscript{(\Letter)}\orcidlink{0000-0002-2812-8781} \and
        Humphrey Shi\inst{4,5}\orcidlink{0000-0002-2922-5663}}

        \footnotetext{\Letter~ Corresponding author.}
        
        \authorrunning{Yihan et al.}
        
        \institute{Institute of Information Science, Beijing Jiaotong University \and
        Visual Intelligence + X International Joint Laboratory of the Ministry of Education \and
        Pengcheng Laboratory, Shenzhen, China \and
        Georgia Institute of Technology \and
        Picsart AI Research (PAIR)\\
        \email{Yihan.hu@bjtu.edu.cn}}

	\maketitle
        \begin{abstract}
Existing natural image matting algorithms inevitably have flaws in their predictions on difficult cases, and their one-step prediction manner cannot further correct these errors. In this paper, we investigate a multi-step iterative approach for the first time to tackle the challenging natural image matting task, and achieve excellent performance by introducing a pixel-level denoising diffusion method (DiffMatte) for the alpha matte refinement. To improve iteration efficiency, we design a lightweight diffusion decoder as the only iterative component to directly denoise the alpha matte, saving the huge computational overhead of repeatedly encoding matting features. We also propose an ameliorated self-aligned strategy to consolidate the performance gains brought about by the iterative diffusion process. This allows the model to adapt to various types of errors by aligning the noisy samples used in training and inference, mitigating performance degradation caused by sampling drift. Extensive experimental results demonstrate that DiffMatte not only reaches the state-of-the-art level on the mainstream Composition-1k test set, surpassing the previous best methods by \emph{\textbf{8\%}} and \emph{\textbf{15\%}} in the SAD metric and MSE metric respectively, but also show stronger generalization ability in other benchmarks. The code will be open-sourced for the following research and applications. Code is available at \href{https://github.com/YihanHu-2022/DiffMatte}{https://github.com/YihanHu-2022/DiffMatte}.


\end{abstract}
	\section{Introduction}

\begin{figure}[t]
    \centering
    \includegraphics[width=0.95\textwidth]{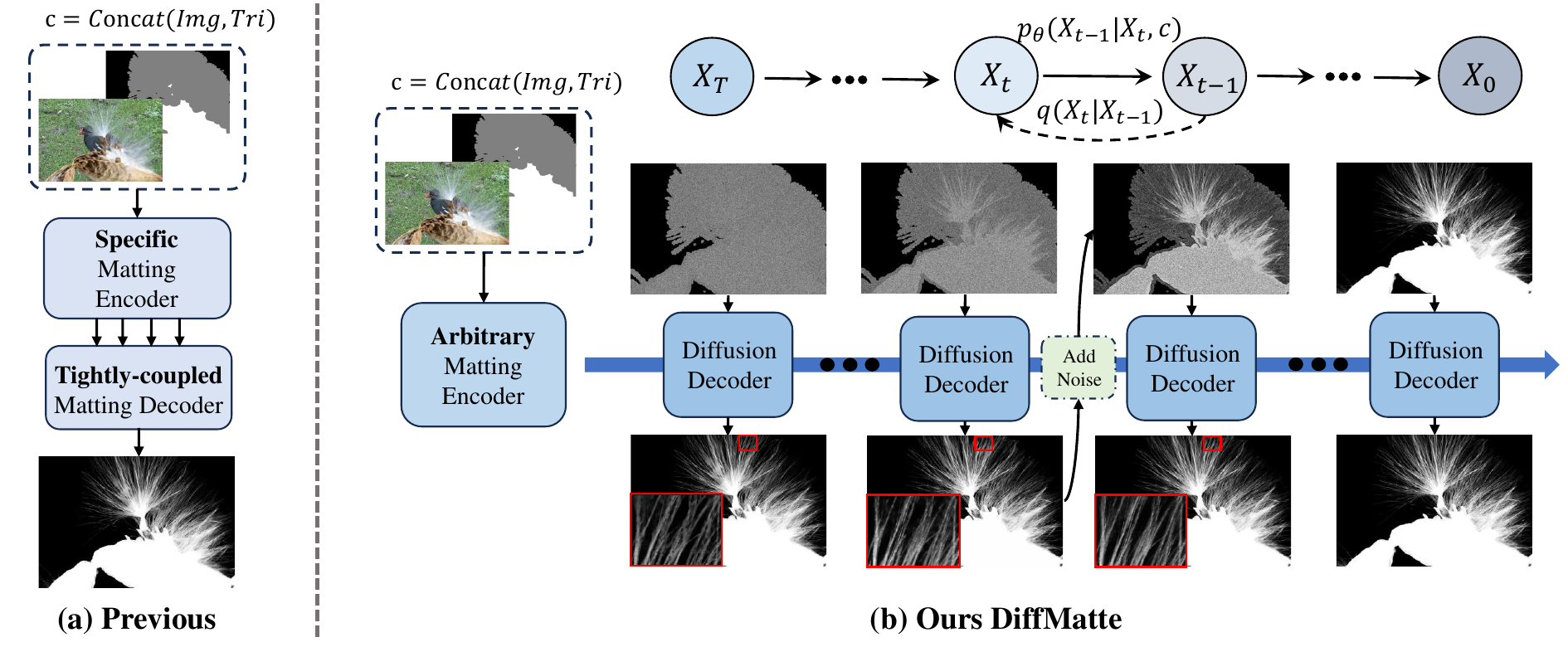}
    \caption{DiffMatte introduces the diffusion process to solve the natural image matting problem. By iteratively correcting the prediction, our method obtains state-of-the-art matting accuracy. DiffMatte can be embedded into arbitrary matting encoders, which makes its application scenarios more flexible and versatile}
    \label{fig:intro}
\end{figure}

Natural image matting is an important task in computer vision, serving the purpose of isolating foreground objects from their backgrounds. Mathematically, a natural image can be expressed as a linear combination of the foreground $F \in \mathbb{R}^{H\times W\times C}$, background $B \in \mathbb{R}^{H \times W \times C}$, and the alpha matte $\alpha \in \mathbb{R}^{H \times W}$, described as:
\begin{equation}
I_{i} = \alpha_{i}F_{i} + (1-\alpha_i)B_{i}, \quad \alpha\in \left [ 0,1 \right],
\end{equation}
Since the foreground color $F_i$, the background color $B_i$, and the alpha value $\alpha$ are left unknown, solving for alpha matte is a highly ill-posed problem. To tackle this, the manually labeled trimaps are used to guide the extraction of foreground opacity with modern deep neural networks~\cite{xu2017deep,lutz2018alphagan,hou2019context,tang2019learning}.

Although the performance of previous one-step matting models keeps on increasing, they still cannot perfectly predict the alpha matte on complex cases, producing conspicuous artifacts or fine area flaws. Considering the success of iterative approaches on segmentation tasks \cite{cheng2020cascadepsp,shen2022high, wang2023segrefiner}, it is an intuitive idea to introduce the iterative mechanism for alpha matte refinement. However, coarse masks are enough to provide semantic priors to guide the iteration of segmentation methods, while matting methods rely on trimap to specifically exploit opacity information in images, which cannot be replaced by coarse alpha mattes. In addition, there is no good way to convert alpha matte into usable trimap. This inconsistency in the form of guidance and prediction prevents the matting task from borrowing the mature iteration framework of the segmentation method, hindering the exploration of iterative alpha matte refinement. 

Recently, the advent of denoising diffusion models \cite{song2019generative,ho2020denoising,song2020denoising} provide an iterative process for high-fidelity and fine-grained generation \cite{nichol2021improved,dhariwal2021diffusion}. We notice that this unique noising and denoising process of the diffusion method naturally forms an iterative paradigm, which avoids the inconsistency between trimap and alpha matte by utilizing the noised alpha matte as additional prior information. In addition, the diffusion method can control this guidance prior by changing the input scaling of noise, and customizing the entire iterative process through the noise schedule. This allows a more flexible utilization of the alpha matte to improve matting quality.

However, applying the diffusion process to natural image matting is non-trivial for the following reasons. First, low iteration efficiency. Existing matting models, to take into account both the specific matting feature and low-level information, usually adopt a tightly coupled network design with specific feature encoders that are bulky in terms of computational overhead. This will lead to redundant calculations while receiving original-sized images (resolution above 2K to ensure clarity) during inference. This difficulty will be further exacerbated by the direct introduction of such a matting model into the iterative process. Second, the performance decline caused by the sampling drift \cite{daras2023consistent}. Due to the recursive nature of the diffusion process, the flawed alpha matte will form the guidance for the next prediction during inference, which deviates from the training samples generated using the ground truth alpha matte. This problem is prominent in matting tasks because the prediction of the alpha matte requires faithful utilization of pixel-level information and demands high accuracy.

To address the challenges that arise when adapting the diffusion process to matting models, we propose the DiffMatte in this paper. Specifically, to reduce the high computational overhead, DiffMatte decouples the image encoder and decoder. As shown in Fig.\ref{fig:intro}, unlike past tightly-coupled matting predictors, diffusion decoder $\mathcal{D}$ only receives the top-level features of arbitrary matting encoder $\mathcal{B}$ without shortcut connections. During the reverse process of inference, only the lightweight $\mathcal{D}$ performs iteratively, and $\mathcal{B}$ acts only once to generate high-dimensional context knowledge, which brings the benefit of a significant reduction in computational overhead. To tackle the sampling drift, DiffMatte includes a modified self-aligned strategy in the later stages of training. We add noise to the alpha matte predicted by the model as a guide to align the samples during training. This helps the model adapt to errors in previous predictions and correctly trade off the prior information for prediction in the current step. Furthermore, our strategy can handle the cumulative effect caused by multi-step errors and maintain stable performance gains during the iteration process.

We perform extensive experiments on a series of composited image matting benchmarks \cite{xu2017deep,qiao2020attention,sun2021semantic} and in-the-wild benchmark AIM-500 \cite{li2021deep} to validate our DiffMatte. When adapted to various matting encoders, DiffMatte outperforms the respective baseline methods on Composition-1k, with the adaptation using ViT-B as the encoder outperforming the previous best method by \textbf{8\%} on the SAD metric (\textbf{18.63}) and by \textbf{15\%} on the MSE metric (\textbf{2.54}). DiffMatte also obtains higher accuracy when generalizing to AIM-500 test sets (SAD \textbf{16.31}, MSE \textbf{3.3}), demonstrating the superior in-the-wild ability of our method. 

\label{sec:intro}
	\section{Related Work}
\label{sec:related}
\noindent\textbf{Natural Image Matting.} Traditional methods are mainly divided into sampling-based~\cite{chuang2001bayesian,wang2007optimized,gastal2010shared,he2011global,shahrian2013improving} and propagation-based methods \cite{sun2004poisson,levin2007closed,levin2008spectral,he2010fast,lee2011nonlocal,chen2013knn}, according to the way they make use of color features. These approaches lack the use of context and prone to producing artifacts. Benefiting from the rapid development of deep learning, learning-based methods can access high-level semantic information with the help of neural networks. \cite{li2020natural,liu2021long,dai2021learning,liu2023rethinking} design learnable modules to exploit contextual knowledge, and \cite{forte2020f,park2022matteformer,dai2022boosting,yu2021high,yao2023vitmatte} introduces stronger backbones \cite{he2016deep,liu2021swin,wang2021pyramid,dosovitskiy2020image} to improve matting accuracy. These methods have made significant progress, but lack exploration of low-level texture features. This leads to matting models relying on shortcut connections to provide low-level features in the one-piece UNet-like structure. In contrast to the above, \cite{yao2023vitmatte} proposes a decoder decoupled architecture to utilize a non-hierarchical backbone network ViT, indicating the unnecessity of previous coupled network design. \cite{li2023matting,yao2023matte} incorporate \cite{kirillov2023segment} to extend matting to any instances.

\noindent\textbf{Diffusion Models.} Diffusion models have achieved significant breakthroughs in various modal generation tasks, owing to their delicate denoising processes. Denoising diffusion probabilistic models (DDPM) \cite{ho2020denoising} accomplish the inverse diffusion process by training a noise predictor for fine-grained image generation. Denoising diffusion implicit models (DDIM) \cite{song2020denoising} use non-Markovian processes to speed up sampling. After some successful attempts \cite{rombach2022high,nichol2021glide} to fuse textual information, a group of generative large models \cite{ramesh2022hierarchical,oppenlaender2022creativity} have achieved surprising results with wide applications in image editing~\cite{liu2023more,goel2023pair,xu2023prompt}. Diffusion models have also been studied for the generation of a wide range of modalities, including video \cite{khachatryan2023text2video,villegas2022phenaki,ho2022imagen,hong2022cogvideo, zhang2023controlvideo}, audio \cite{kong2020diffwave,kolesnikov2020big,jeong2023power}, biomedical image \cite{trippe2022diffusion,schneuing2022structure} and text \cite{li2022diffusion,gong2022diffuseq}.

\noindent\textbf{Diffusion models for perception tasks.} Diffusion methods attract extensive research interest due to the success of diffusion modeling in the generative field. Since the pioneering work \cite{amit2021segdiff} introduced diffusion methods to solve image segmentation, follow-up researchers use diffusion to attempt their respective tasks. \cite{chen2023diffusiondet} formulates object detection as a denoising process. \cite{saxena2023monocular} involves a diffusion pipeline into depth estimation approach. \cite{gu2022diffusioninst}, \cite{chen2023generalist}, and \cite{lai2023denoising} apply diffusion to instance segmentation, panoptic segmentation, and semantic segmentation respectively, where the diffusion denoising training technique used by Pix2Seq \cite{chen2023generalist} is utilized by DDP \cite{ji2023ddp} to solve diversified dense prediction tasks. These works have achieved good performance by introducing the diffusion process, but \cite{saxena2023monocular,chen2023diffusiondet,ji2023ddp} observe that the performance decreases as the step increases, and this phenomenon is exacerbated on dense prediction tasks with high accuracy demand.
	\section{DiffMatte}
\label{sec:model}


\begin{figure}
    \centering
    \includegraphics[width=0.90\textwidth]{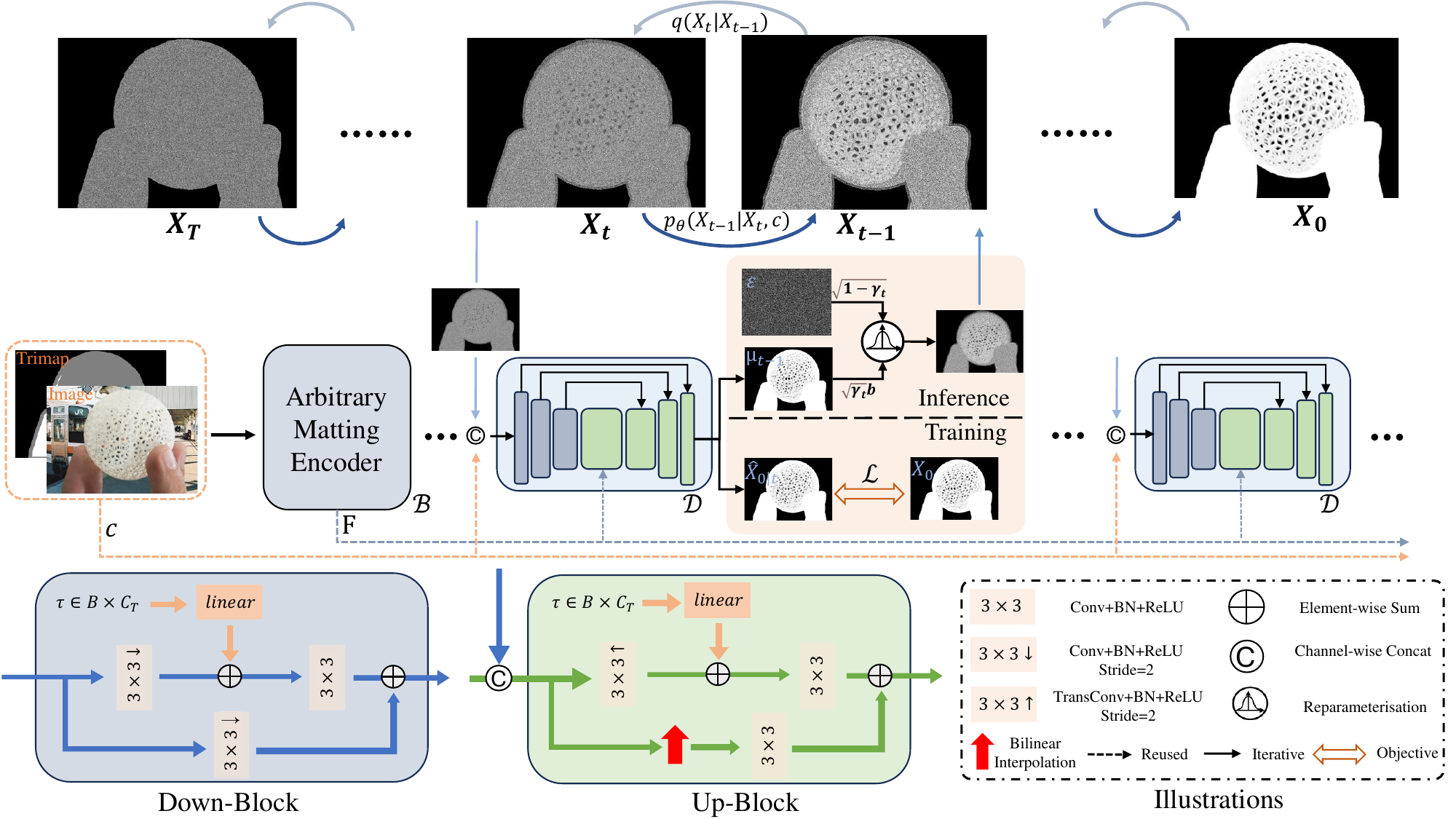}
    \caption{\textbf{Proposed DiffMatte Framework.} DiffMatte provides noise sample $X$ for training through the forward diffusion process $q$, and uses X as an iterative prior to supplement the guidance of $c$ in the reverse process $p_\theta$. Diffusion decoder $\mathcal{D}$ serves as an iterative component and is independent of matting encoder$\mathcal{B}$, receiving matting features $F$ to predict the alpha matte at each step. As a universal decoder, it can cooperate with any matting encoder. $\mathcal{D}$ consists of Down-Block and Up-Block, which only contain convolutional layers and linear layers to process image information and time embedding $\tau$ respectively.}
    \label{fig:pipeline}
\end{figure}

\subsection{Constructing DiffMatte Framework}
In this section, we introduce our iterative matting framework with the task-specific diffusion process. Each step of DiffMatte's prediction is based on the fixed image and trimap $c=Concat(Image, Trimap)$, and accepts noise sample $X$ as additional guidance. The iteration process starts with pure Gaussian noise and ends up converging to a clean alpha matte. This gradual change enables the model to balance the guidance of $c$ and $X$ during the process.\\
\noindent\textbf{DiffMatte Framework.} To iteratively refine the alpha matte, we carefully design DiffMatte’s framework. The key point of this framework is the diffusion decoder $\mathcal{D}$, which is the only unit of the iterative process. $\mathcal{D}$ receives $X_t$ at each step $t$ as an additional pixel-level prior. This prior information needs to be fused with the matting features provided by the matte encoder $\mathcal{B}$ to generate the alpha matte prediction of the current step. We design $\mathcal{D}$ as a symmetric UNet-like network to complete this process. This iterative $\mathcal{D}$ can cooperate with any matting encoder $B$, forming a general matting architecture.

\noindent\textbf{Providing Training Samples through Forward Process.} Different from the previous one-step matting method, DiffMatte needs to obtain noise samples during training for each iteration, which is completed through the forward process in the diffusion method. We adopt the following equation \cite{chen2023importance} to define the forward process:
\begin{equation}
     X_t \sim q(X_t|X_0) = \sqrt{\gamma_t}(bX_0) + \sqrt{1 - \gamma_t}\epsilon
     \label{eq:2}
\end{equation}
where $\gamma_t \in (0, 1)$ is a mapping of $t$ through the noisy schedule and represents the noise intensity. $X_0$ indicates the ideal clean sample of alpha matte. $\epsilon$ is a standard Gaussian noise. In the previous research, three noisy schedules are set up for image generation but also work for perception tasks, namely linear schedule \cite{ho2020denoising,chen2023importance}, cosine schedule \cite{nichol2021improved}, and sigmoid schedule \cite{jabri2022scalable}. Our experiments show that simple linear schedules are more suitable for matting. $b \in \left( 0, 1 \right ]$ denotes the input scaling factor, which amplifies the noises. A smaller $b$ will bring more destruction of detailed information at the same $\gamma_t$, which can be interpreted as an increase to the signal-to-noise ratio (SNR) \cite{chen2023generalist,chen2023importance}. 

In the current phase of training, we sample a single time step $t$ from a uniform distribution $U(0, 1)$ following the continuous time training paradigm \cite{kingma2021variational,chen2022analog}, and noise $X_0$ to $X_t$ for an iteration training according to Eq. \ref{eq:2}.

\noindent\textbf{Iterative Prediction in Reverse Process.} Given the noisy sample $X_t$, DiffMatte obtains the denoised sample $X_{t-1}$ through estimating $\mu_{t-1} = \hat{X}_{0|t}$ with trained $f_\theta$. After that a sampling technique proposed by DDIM~\cite{song2020denoising} is used to sample $X_{t-1}$, which can be defined as:
\begin{equation}
     X_{t-1} \sim \mathcal{N}(X_{t-1}; \sqrt{\gamma_{t-1}}\mu_{t-1}, (1 - \gamma_{t-1})\textbf{I})
\end{equation}
Combining the estimation of $\mu_{t-1}$ and DDIM sampling, we get one iteration of the reverse process:
\begin{equation}
     p_{\theta}(X_{t-1}|X_t, c) = \mathcal{N}(X_{t-1};\sqrt{\gamma_{t-1}}f_{\theta}(X_t, t, c), (1-\gamma_{t-1})\textbf{I})
\end{equation}
The complete reverse process starts with a standard Gaussian noise $X_T$ and passes through $T$-step iteration to get the final estimation $\hat{X}_0$. During inference, as the time step goes from $T$ to $0$, the value of $\gamma_t$ is mapped from $1$ to $0$ via the noisy function. Thus $X_T \sim \mathcal{N}(0, I)$, which is consistent with $\epsilon$ at the beginning time step $T$.

\subsection{Iterative Refinement}
\label{sec:iterative}
\noindent\textbf{Efficiency.} The network designed for the matting task is usually a tightly coupled one-piece structure \cite{hou2019context,li2020natural,dai2021learning,dai2022boosting}, joining the entire network to the iterative diffusion process without modification leads to excessive computational overhead. Inspired by structural designs of \cite{yao2023vitmatte}, we decouple the network into separate matting encoder $\mathcal{B}$ and iterative diffusion decoder $\mathcal{D}$, and the connection between the two is limited to the top-level context feature of $\mathcal{B}$. We implement the DiffMatte model $f_\theta$ as:
\begin{equation}
    \begin{split}
     & F = \mathcal{B}(c)  \\
     f_{\theta}(X_t, t, c) = \mathcal{D}&(cat(X_t, c), t, F) = \mu_{t-1}
     \end{split}
\end{equation}

\begin{table}[t]
    \centering
    \setlength{\tabcolsep}{2.2mm}
    \renewcommand{\arraystretch}{1.1}{
    \caption{\textbf{Computational overhead and running time.} We use an image with a resolution of $2048\times2048$ as input, with computational overhead in TFLOPs. SAD$_1$ indicates one-step prediction results on Composition-1k benchmark. $first$ and $sub$ represents the computation and time consumption generated by the first step and subsequent iterations respectively.}
    \label{tab:costs}
    \begin{tabular}{l|ccccc}
        \toprule
        Models & TFLOPs$_{first}$ & TFLOPs$_{sub}$  & Times$_{first}$ & Times$_{sub}$ & SAD$_1$\\
        \hline
        GCA~\cite{li2020natural} & 2.09 & - & 357$_{ms}$ & - & 35.3 \\
          \rowcolor{gray!15}DiffMatte-Res34 & 1.12 & 0.86$_{\textcolor{red}{~24\%\downarrow}}$ & 156$_{ms}$ & 82$_{ms \textcolor{red}{~47\%\downarrow}}$ & \textbf{31.3} \\
        \hline
        ViTMatte-S~\cite{yao2023vitmatte} & 1.69 & - & 617$_{ms}$ & - & 21.46 \\
         \rowcolor{gray!15} DiffMatte-ViTS & 2.08 & 0.82$_{\textcolor{red}{~60\%\downarrow}}$ & 578$_{ms}$ & 126$_{ms \textcolor{red}{~78\%\downarrow}}$ & \textbf{20.61} \\
        \bottomrule
    \end{tabular}}
\end{table}

As shown in Figure \ref{fig:pipeline}, $\mathcal{B}$ encodes the image with the information of trimap to get the context knowledge $F$, which will be reused in the reverse process. $\mathcal{D}$ is iteratively performed in the diffusion process, and its lightweight structure prevents excessive computational overheads. We construct the decoupled diffusion decoder $\mathcal{D}$ with Down-Block and Up-Block. They are essentially residual convolution modules that incorporate time step information $\tau$ encoded by linear mapping of time step $t$. Each module contains only 3 convolutional layers and a linear layer encoding temporal embedding. It uses only one pair of blocks per feature resolution instead of repeated stacking, making it very lightweight and fast (5.3M parameters and 126$ms$ running time). More network details can be found in the appendix.

\setlength{\intextsep}{-3pt}
\begin{wrapfigure}{r}{0.48\textwidth}
    \centering
    \includegraphics[width=0.47\textwidth]{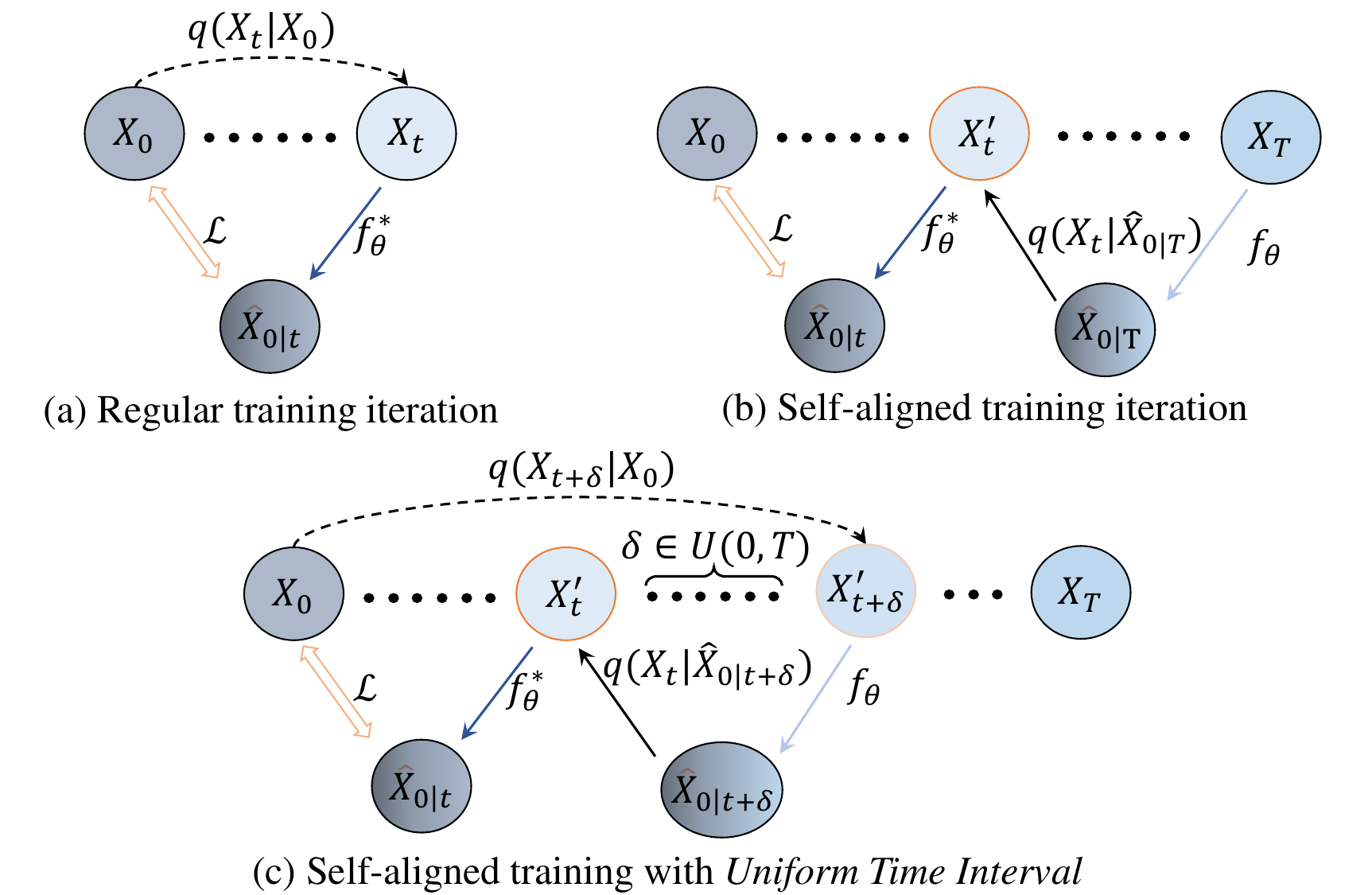}
    \caption{Comparison of different training strategies. In contrast to the strategy proposed by \cite{ji2023ddp}, we take into account the drift caused by prediction errors at each time step and propose a self-aligned strategy with uniform time intervals to align the noisy sample $X$ over all time steps.}
    \label{fig:UD}
\end{wrapfigure}

In this way, the diffusion process can be smoothly implemented using the lightweight diffusion decoder, and the heavy encoder only propagates once to avoid computational redundancy. As shown in Table \ref{tab:costs}, with this lightweight design, DiffMatte can save 24\% of the computational overhead and 47\% of the inference time in each subsequent iteration when using the ResNet34 encoder, and can save up to 60\% of the computational overhead and 78\% of the inference time when using the ViTS encoder. Compared to directly using the entire matting network for iteration, DiffMatte utilizes computing resources more efficiently and significantly reduces inference time.


\noindent\textbf{Self-aligned Strategy with Uniform Time Intervals.} The iterative process provided by the diffusion method is supposed to acquire performance gains with the step growth, but the results abnormally exhibit a continuous performance degradation (the evidence is displayed in Tab.\ref{tab:nands}). This phenomenon can also be found in perceptual methods that introduce diffusion processes \cite{saxena2023monocular,chen2023diffusiondet,ji2023ddp}. 

We attribute this phenomenon to the inconsistency of noise samples between training and inference. If only $X$ formed by ground-truth alpha matte is used for guidance during training, the model will not be able to adapt to error-containing $X$ during inference. Especially with the recursive nature of the iterative manner, errors in $X$ will accumulate, exacerbating the loss of performance. This view is similarly mentioned in DDP \cite{ji2023ddp}, but the solution it proposes is limited as the ignorance of the accumulation of errors that occurs during the iterative process. 

We propose a more refined strategy that can effectively correct this distribution variance by converting sampling targets to time intervals, obtaining a self-aligned strategy with Uniform Time Intervals (UTI). As shown in Fig.\ref{fig:UD}, we sample a time interval $\delta \in U(0, T)$, which acts on the training time step $t \in U(0, T - \delta)$ to obtain $\acute{t} = t + \delta$. In the subsequent procedures, we use an additional forward process $q(X_{\acute{t}}|X_{0})$ to calculate the previous step noise sample instead of using white noise like \cite{ji2023ddp}. After that, we obtain the estimation $\hat{X}_{0|t+\delta}$ of the current sample $X^{'}_{t+\delta}$ over $X_0$ by a frozen diffusion model $f_{\theta}$. After replacing $X_0$ with $\hat{X}_{0|t+\delta}$ we perform a regular training iteration. The use of $X_{\acute{t}}$ helps the model to complete the alignment of the data distribution over the entire time domain and mitigate the accumulation of errors. Once $\delta$ takes the value of $0$, our method reverts to regular training, while when $\delta$ takes the value of $T - t$ it becomes the case used by \cite{ji2023ddp}. We add our UTI self-aligned strategy after $f_{\theta}$ is well-trained to prevent serious errors in estimation $\hat{X}_{0|\acute{t}}$ from causing the training failure. 

\noindent
\begin{minipage}{0.49\textwidth}
\begin{algorithm}[H]
\caption{\scriptsize DiffMatte Training}
\label{algo:train}
\definecolor{codeblue}{HTML}{2E8B57} 
\definecolor{codekw}{HTML}{DC143C} 
\lstset{
  backgroundcolor=\color{white},
  breaklines=true,
  captionpos=b,
  commentstyle=\fontsize{12pt}{12pt}\color{codeblue},
  keywordstyle=\fontsize{12pt}{12pt}\color{codekw},
  escapechar={|}, 
}
\lstset{language=Python}
\begin{lstlisting}[xleftmargin=-2.5em]
def train(cond, pha, flag, b):
    """  cond: [B, 4, H, W], pha: [B, 1, H, W] """
    """ flag: self-align entry, b: input scale """
    feat = mat_encoder(cond) # encode condition
    pha = (pha * 2) - 1 # normalize
    # forward process
    t, eps = uniform(0, 1), normal(0, 1)
    if flag == True:
        Xt = self_align(t, eps) # get aligned sample
    else:
        gamma = noisy_func(t)
        Xt = sqrt(gamma) * pha + sqrt(1-gamma) * eps
    # predict and backward
    X0 = diff_decoder(Xt, cond, feat, t)
    X0 = (X0 + 1) / 2
    loss = Losses(X0, pha)
    return loss
\end{lstlisting}
\end{algorithm}
\end{minipage}
\hfill
\begin{minipage}{0.47\textwidth}
\begin{algorithm}[H]
\caption{\scriptsize DiffMatte Inference}
\label{algo:inference}
\definecolor{codeblue}{HTML}{2E8B57} 
\definecolor{codekw}{HTML}{DC143C} 
\lstset{
  backgroundcolor=\color{white},
  breaklines=true,
  captionpos=b,
  commentstyle=\fontsize{12pt}{12pt}\color{codeblue},
  keywordstyle=\fontsize{12pt}{12pt}\color{codekw},
  escapechar={|}, 
}
\lstset{language=Python}
\begin{lstlisting}[xleftmargin=-2.5em]
def inference(cond, T, b):
    """ cond: [B, 4, H, W], T: sampling steps """
    """       b: input scale        """
    feat = mat_encoder(cond) # encode condition
    Xt = normal(0, 1) # noisy map of [B, 1, H, W]
    time_pairs = sample_timesteps(T)
    # reverse process
    for t, t_next in time_pairs:
        gamma, gamma_next = noisy_func(t, t_next)
        # normalize X_t by variance
        Xt = Xt / std(Xt)
        # predict X0
        X0 = diff_decoder(Xt, cond, feat, t)
        X0 = (X0 * 2) - 1 # normalize
        Xt = DDIM(X0, gamma, gamma_next, b)
    Xpred = Xt / b # rescaling
    return [Xpred + 1] / 2 # denormalize
\end{lstlisting}
\end{algorithm}
\end{minipage}

\subsection{Training and Inference.} 

Our training and inference algorithms are shown in Algorithm \ref{algo:train} and Algorithm \ref{algo:inference}. Our approach requires a selected noise schedule as well as input scaling, both of which parameterize the corresponding noise distribution, leading to different diffusion processes. DiffMatte involves the timing of the start of the UTI self-aligned strategy during training. We train to the 90th epoch to add it and continue until the end of training. At training time our $f_{\theta}$ is supervised with the task-specific losses following the common practices \cite{chen2023generalist,chen2023diffusiondet,ji2023ddp}:
\begin{equation}
     \mathcal{L} = E_{t \sim U(1, T), X_t \sim q(X_t|X_0,\textbf{I})} \mathcal{L}_{mat}(f_{\theta}(X_t, t, c), X_0)
\end{equation}

specifically using the combined matting loss with separate $l1$ loss \cite{yao2023vitmatte}, $l2$ loss, laplacian loss \cite{hou2019context}, and gradient penalty loss. We end up with the following objective:
\begin{equation}
     \mathcal{L}_{mat} = \mathcal{L}_{sp \ l_1} + \mathcal{L}_{l_2}+ \mathcal{L}_{lap} + \mathcal{L}_{grad}
\end{equation}

Thanks to continuous-time training, we are free to set the total number of iterations $T$ during inference. As demonstrated in Fig.\ref{fig:c1k_iter}, an increase in the number of sample steps from 1 to 10 is accompanied by an improvement in the accuracy of the final prediction.
	\section{Experiments}
\label{sec:experiment}

\subsection{Datasets and Evaluation.} 
\noindent\textbf{Adobe Image Matting \cite{xu2017deep}.}  This dataset contains 431 unique training foreground images and 50 extra foregrounds for
evaluation. The training set is constructed by compositing each foreground with 100 background images from the COCO dataset \cite{lin2014microsoft}. Similarly, the validation set named Composition-1k is obtained by compositing 50 test foreground images with 20 background images from VOC2012 \cite{everingham2010pascal} to get a total of 1000 test images. 

\setlength{\intextsep}{-3pt}
\begin{wraptable}[27]{r}{0.6\textwidth}
    \centering
    \footnotesize
    \setlength{\tabcolsep}{0.2pt}
    \caption{Quantitative results on Composition-1k ~\cite{xu2017deep}. $^\dagger$ indicates using the perturbation mask as guidance. The best results are shown in bold. S1 and S10 denote 1 and 10 steps refinement.}
    \label{tab:composition1k}
    \begin{tabular}{l|cccc}
        \toprule
        Methods & SAD & MSE($10^3$) & Grad & Conn \\
        \hline
        DIM~\cite{xu2017deep} & 50.4 & 14.0 & 31.0 & 50.8 \\
        IndexNet~\cite{lu2019indices} & 45.8 & 13.0 & 25.9 & 43.7 \\
        SampleNet~\cite{tang2019learning} & 40.4 & 9.9 & - & - \\
        Context-Aware~\cite{hou2019context} & 35.8 & 8.2 & 17.3 & 33.2 \\
        A$^2$U~\cite{dai2021learning} & 32.2 & 8.2 & 16.4 & 29.3 \\
        MG$^\dagger$~\cite{yu2021mask} & 31.5 & 6.8 & 13.5 & 27.3 \\
        SIM~\cite{sun2021semantic} & 28.0 & 5.8 & 10.8 & 24.8 \\
        FBA~\cite{forte2020f} & 25.8 & 5.2 & 10.6 & 20.8 \\
        TransMatting~\cite{cai2022transmatting} & 24.96 & 4.58 & 9.72 & 20.16 \\
        RMat~\cite{dai2022boosting} & 22.87 & 3.9 & 7.74 & 17.84 \\
        \hline
        GCA~\cite{li2020natural} & 35.3 & 9.1 & 16.9 & 32.5 \\
         \rowcolor{gray!15} DiffMatte-Res34 (S1) & 31.28 & 6.38 & 11.60 & 28.07 \\
         \rowcolor{gray!15} DiffMatte-Res34 (S10) & \textbf{29.20} & \textbf{6.04} & \textbf{11.31} & \textbf{25.48} \\
        \hline
        Matteformer~\cite{park2022matteformer} & 23.80 & 4.03 & 8.68 & 18.90 \\
         \rowcolor{gray!15} DiffMatte-SwinT (S1) & 22.05 & 3.54 & 6.67 & 17.03 \\
         \rowcolor{gray!15} DiffMatte-SwinT (S10) & \textbf{20.87} & \textbf{3.23} & \textbf{6.37} & \textbf{15.84} \\
        \hline
        ViTMatte-S~\cite{yao2023vitmatte} & 21.46 & 3.3 & 7.24 & 16.21 \\
         \rowcolor{gray!15} DiffMatte-ViTS (S1) & 20.61 & 3.08 & 7.14 & 14.98 \\
         \rowcolor{gray!15} DiffMatte-ViTS (S10) & \textbf{20.52} & \textbf{3.06} & \textbf{7.05} & \textbf{14.85} \\
        \hline
        ViTMatte-B~\cite{yao2023vitmatte} & 20.33 & 3.0 & 6.74 & 14.78 \\
         \rowcolor{gray!15} DiffMatte-ViTB (S1) & 18.84 & 2.56 & 5.86 & 13.23 \\
         \rowcolor{gray!15} DiffMatte-ViTB (S10) & \textbf{18.63} & \textbf{2.54} & \textbf{5.82} & \textbf{13.10} \\
        \bottomrule
    \end{tabular}
\end{wraptable}

\noindent\textbf{Generalization.} We use the test sets of Distinctions-646 \cite{qiao2020attention} (hereinafter D646) and Semantic Image Matting \cite{sun2021semantic} (hereinafter SIMD) to verify the generalization performance of DiffMatte. D646 and SIMD contain 50 and 39 test foregrounds respectively, which are composited with the background in Pascol-VOC to obtain test images. SIMD provides the trimap of the foreground, while the trimap of D646 needs to be generated by ourselves.

\noindent\textbf{AIM-500 \cite{li2021deep}.} AIM-500 is the most comprehensive real image test set among the natural image matting benchmarks. It contains 500 real images with official trimap and detailed alpha matte annotations. We choose AIM-500 to evaluate DiffMatte's in-the-wild performance.

We train our model on the Adobe Image Matting training set and perform inference on three composite image test sets to validate the matting performance as well as the generalization. We additionally use the training set in D646 for DiffMatte training of the ViT series and verify it on real-world images. The reason for using the D646 training set is that it has a wider data domain, containing more natural categories including flames and liquids. We use 4 common metrics to evaluate our DiffMatte: Sum of Absolute Differences (\textbf{SAD}), Mean Square Error (\textbf{MSE}), Gradient loss (\textbf{Grad}), and Connectivity loss (\textbf{Conn}). A lower value indicates better quality. 

\begin{figure*}
    \centering
    \includegraphics[width=0.98\textwidth]{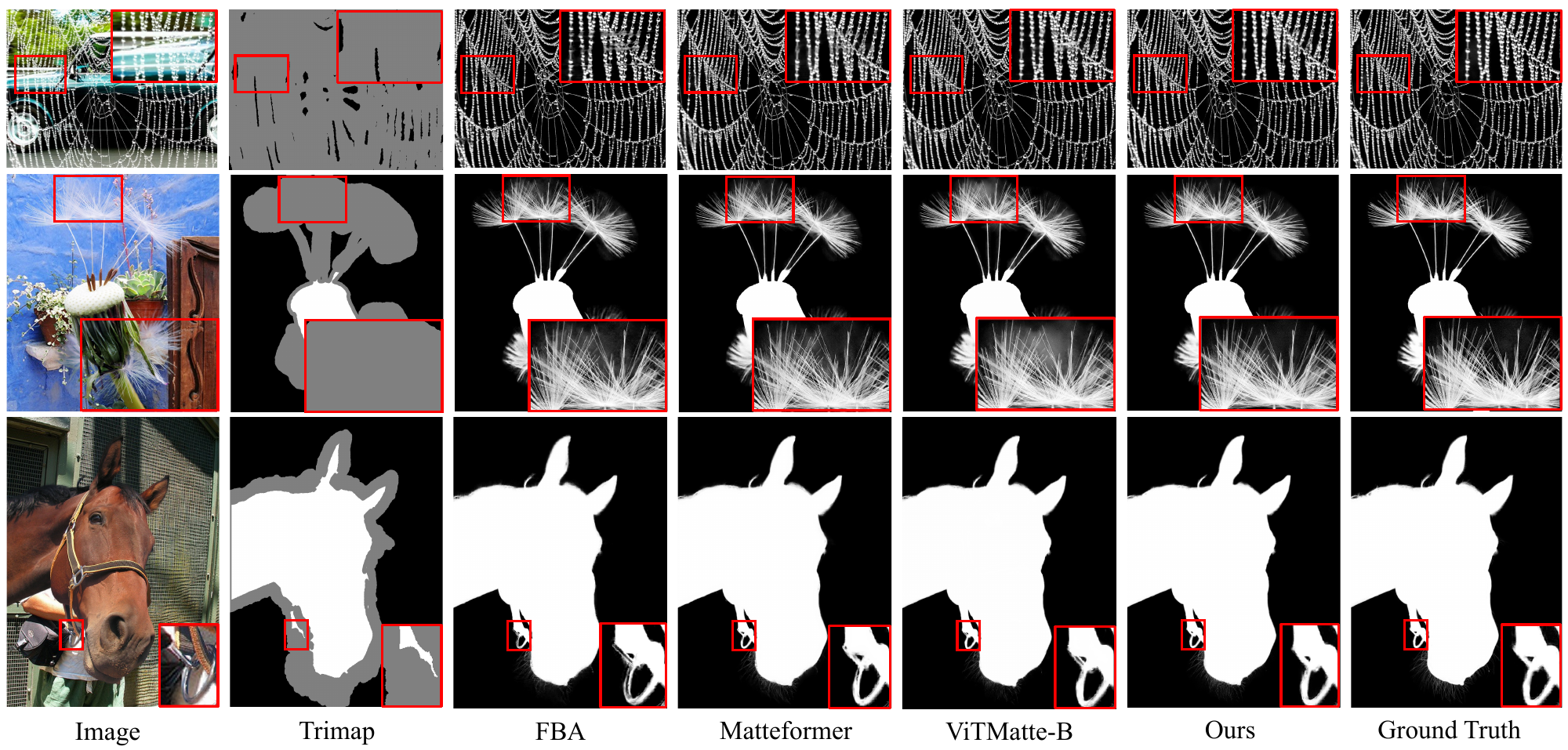}
    \caption{Qualitative results compared with previous SOTA methods on Composition-1k. }
    \label{fig:c1k}
\end{figure*}

\subsection{Main Results}
\label{sec:result}
In this section, We deploy DiffMatte's general decoder on four popular matting encoders \cite{li2020natural,park2022matteformer,yao2023matte} and present our quantitative and qualitative comparison results with previous methods on various benchmarks. Then we discuss the performance improvements brought by DiffMatte's unique iteration manner.

\setlength{\intextsep}{-3pt}
\begin{wrapfigure}{r}{0.42\textwidth}
    \centering
    \includegraphics[width=0.4\textwidth]{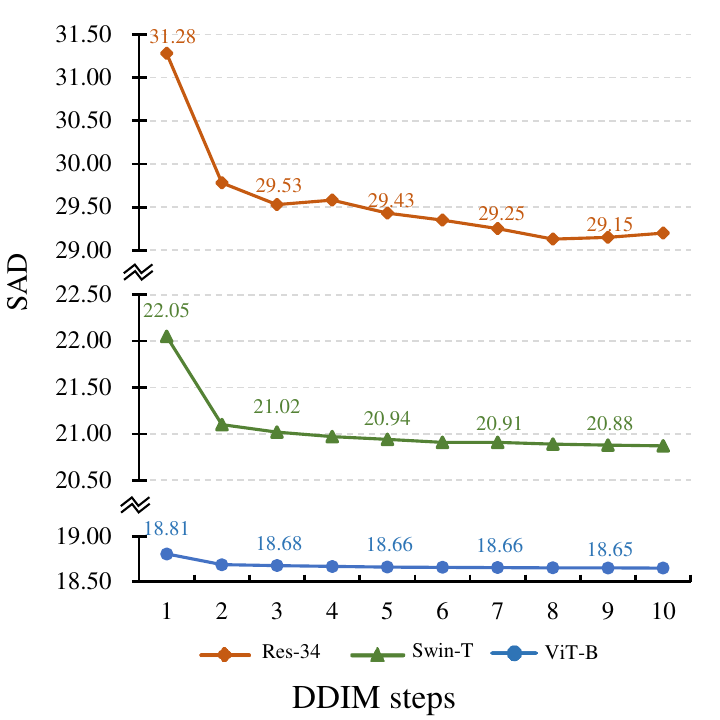}
    \caption{The iteration results of each step of DiffMatte with different matting encoders.}
    \label{fig:c1k_iter}
\end{wrapfigure}
\noindent\textbf{Results on Composition-1k.} The quantitative results on Composition-1k are shown in Tab.\ref{tab:composition1k}. With the ViT-B encoder, one-step DiffMatte achieves the best matting accuracy, improving the SAD metric by 1.49 (+7.3\%) and the MSE metric by 0.44 (+14.7\%) compared with the previous SOTA method . Fig.\ref{fig:c1k_iter} shows the refining process of each step on the Composition-1k with different matting encoders. The overall results improve as the number of iteration steps increases. Variants with Swin Tiny and ResNet-34 encoders enjoy more significant improvement. We will further explain this phenomenon in the following subsection. Fig.\ref{fig:c1k} qualitatively comparing our approach to previous SOTA methods, represents the better performance in the challenging local regions, demonstrating the superiority of DiffMatte. 

\begin{table}[t!]
    \centering
    \scriptsize
    \setlength\tabcolsep{2.7 pt}
    \renewcommand\arraystretch{1.4} 
    \caption{Quantitative results on D646~\cite{qiao2020attention} and SIMD~\cite{sun2021semantic}. All methods are trained only on the Adobe Image Matting dataset. S1 and S10 denote 1 and 10 steps refinement.}
    \label{tab:adaption}
    \scalebox{0.8}{
    \begin{tabularx}{\textwidth}{l|rrrr|m{0.88cm}<{\centering}m{0.88cm}<{\centering}m{0.88cm}<{\centering}m{0.88cm}<{\centering}|c}
        \toprule
         Dataset & \multicolumn{4}{c|}{Distinctions-646} & \multicolumn{4}{c|}{Semantic Image Matting Dataset} & \multirowcell{2}{Params$\;\;$}\\
         Method & SAD & MSE & Grad & Conn & SAD & MSE & Grad & Conn & \\
        \midrule
         GCA~\cite{li2020natural} & 35.33 & 18.4 & 28.78 & 34.29 & 68.23 & 25.74 & 33.19 & 67.67 & 25.3M$\quad$\\
         \rowcolor{gray!15} DiffMatte-Res34 (S1) & 31.53 & 11.87 & 17.52 & 30.86 & 51.49 & 14.10 & 18.32 & 48.70 & \\
         \rowcolor{gray!15} DiffMatte-Res34 (S10) & \textbf{29.38} & \textbf{11.31} & \textbf{16.19} & \textbf{28.26}  & \textbf{47.75} & \textbf{14.10} & \textbf{17.19} & \textbf{44.53} & \multirowcell{-2}{23.9M$\quad$}\\
         \midrule
         Matteformer~\cite{park2022matteformer} & 23.90 & 8.16 & 12.65 & \textbf{18.90} & 29.66 & 5.91 & 12.52 & 24.19 & 48.8M$\quad$\\
         \rowcolor{gray!15} DiffMatte-SwinT (S1) & 23.46 & 6.71 & 10.20 & 21.23 & 30.26 & 5.64 & 9.45 & 24.64 & \\
         \rowcolor{gray!15} DiffMatte-SwinT (S10) & \textbf{23.17} & \textbf{6.58} & \textbf{10.04} & 20.03 & \textbf{27.51} & \textbf{5.20} & \textbf{9.12} & \textbf{22.04} & \multirowcell{-2}{38.8M$\quad$}\\
         \midrule
         ViTMatte-S~\cite{yao2023vitmatte} & 23.18 & 7.14 & 13.97 & 19.65 & 27.96 & 5.02 & 10.68 & 22.38 & 25.8M$\quad$\\
         \rowcolor{gray!15} DiffMatte-ViTS (S1) & \textbf{22.56} & \textbf{7.09} & 13.21 & \textbf{19.23} & 27.55 & 4.78 & \textbf{10.26} & 21.24  &\\
         \rowcolor{gray!15} DiffMatte-ViTS (S10) & 22.96 & 7.22 & \textbf{13.06} & 19.66 & \textbf{27.38} & \textbf{4.71} & 10.31 & \textbf{21.03} &\multirowcell{-2}{29.0M$\quad$} \\
         \midrule
         ViTMatte-B~\cite{yao2023vitmatte} & 20.36 & 5.58 & 9.34 & 17.19 & 27.15 & 5.45 & 9.67 & 21.51 & 96.7M$\quad$\\
         \rowcolor{gray!15} DiffMatte-ViTB (S1) & \textbf{19.07} & \textbf{5.23} & 9.26 & \textbf{15.99} & 26.83 & 4.92 & 8.26 & 20.72 & \\
         \rowcolor{gray!15} DiffMatte-ViTB (S10) & 19.19 & 5.34 & \textbf{9.26} & 16.17 & \textbf{25.60} & \textbf{4.69} & \textbf{8.20} & \textbf{19.84} &\multirowcell{-2}{101.4M$\ \;$}\\
        \bottomrule
    \end{tabularx}}
\end{table}

\noindent\textbf{Generalization on D646 and SIMD.} The quantitative results on the D646 and SIMD test sets are shown in Tab.\ref{tab:adaption}. All baselines use the official weights trained on the Adobe Image Matting training set to test generalizability. The results indicate that DiffMatte outperforms the competitors on both the D646 and SIMD test sets. Notably, DiffMatte based on the ViT backbone shows a trend of performance degradation with increasing time steps on the D646 test set. We speculate that this is due to the low feature resolution provided by ViT (16x downsampling), resulting in a shallow supporting decoder that is insufficient for handling the D646 test set, which significantly differs from the training domain. We additionally trained the ViTS and ViTB models on the D646 training set, and both recovered their refinement capabilities after training on a larger data domain. (Results shown in the Appendix)

\begin{figure*}
    \centering
    \includegraphics[width=0.94\textwidth]{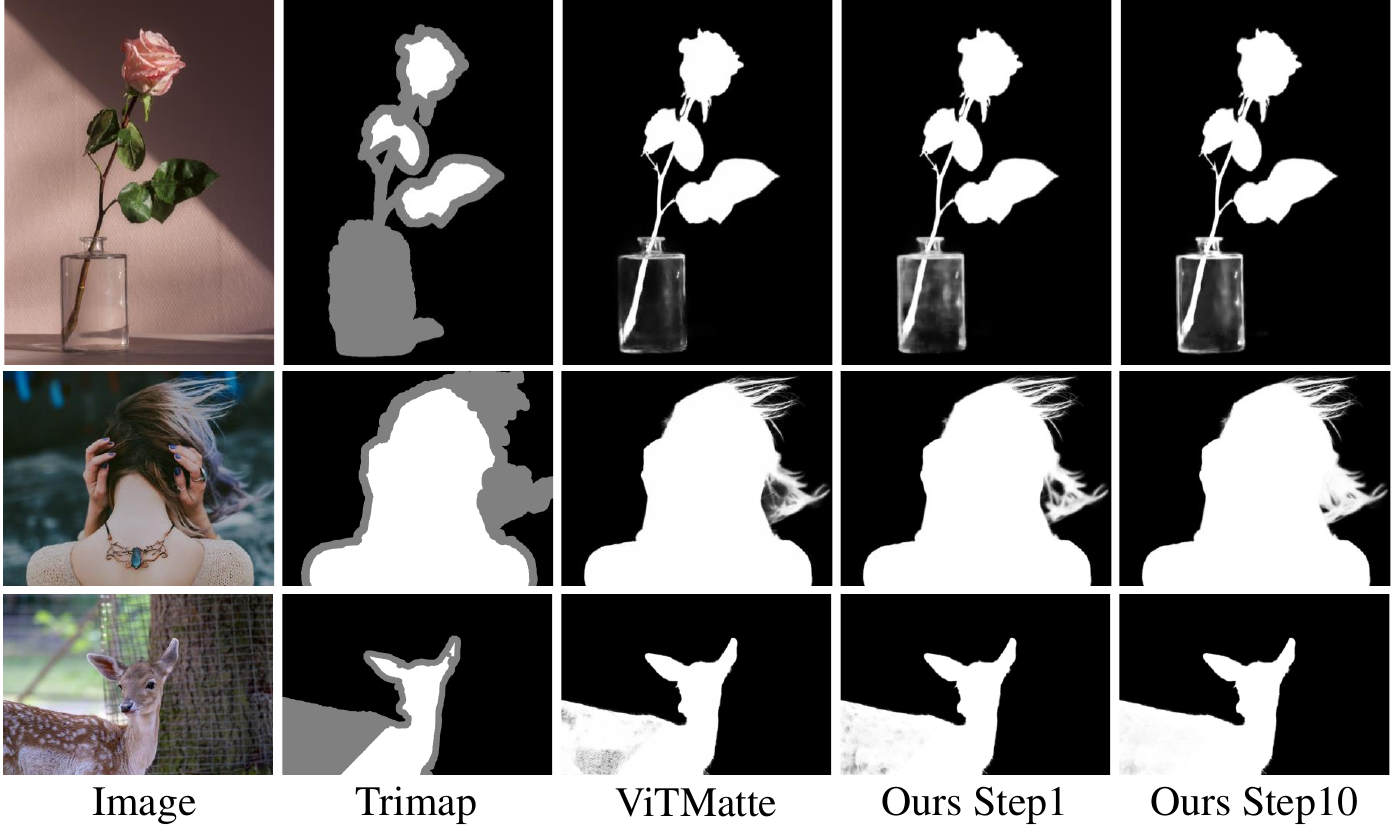}
    \caption{We select examples to represent three difficult natural image matting scenarios. The first row is the scene with complex foregrounds containing fully transparent objects. The second row is a scene with a complex background, including strong light and blur. The third row is the scene with only a rough trimap for guidance. }
    \label{fig:AIM}
\end{figure*}

\setlength{\intextsep}{-3pt}
\begin{wraptable}[11]{r}{0.6\textwidth}
    \centering
    \footnotesize
    \setlength{\tabcolsep}{0.2pt}
    \caption{Quantitative results on AIM-500 ~\cite{li2021deep}. $^\ddagger$ indicates training on Distinctions-646. The best results are shown in bold. S1 and S10 denote 1 and 10 steps refinement.}
    \label{tab:AIM}
    \begin{tabular}{l|cccc}
        \toprule
        Methods & SAD & MSE($10^3$) & Grad & Conn \\
        \hline
         ViTMatte-B~\cite{yao2023vitmatte} & 17.93 & 1.88 & 15.52 & 17.2 \\
         \hline
         DiffMatte-ViTB (S1) & 22.32 & 1.9 & 14.63 & 22 \\
         DiffMatte-ViTB (S10) & 23.57 & 2.14 & \textbf{14.58} & 23.23 \\
         DiffMatte-ViTB$^\ddagger$ (S1) & 17.06 & 1.73 & 15.39 & 16.87 \\
         DiffMatte-ViTB$^\ddagger$ (S10) & \textbf{16.73} & \textbf{1.7} & 14.78 & \textbf{16.35} \\
        \bottomrule
    \end{tabular}
\end{wraptable}
\noindent\textbf{Results on AIM-500.} We compare the results of training on the Adobe Image Matting training set and Distinction-646 training set based on ViT-B encoder with the previous strongest model, ViTMatte, on the in-the-wild benchmark AIM-500. The results are shown in the Tab.\ref{tab:AIM} . We still provide the results of running one step and iterating ten steps respectively. We find that the effect of DiffMatte in Distinction-646 training is significantly better than Adobe Image Matting, and the effect will not deteriorate with iteration. This phenomenon shows that training on synthetic images using a wider data domain can help improve DiffMatte’s generalization ability in the real world. DiffMatte's performance is better than ViTMatte in all four metrics, reflecting its effectiveness on real images. We further provide visualization results in Fig.\ref{fig:AIM}, from which we can see the quality improvements through multi-step iteration.

\subsection{Discussion} 
\noindent\textbf{Promising one-step results.} When DiffMatte is applied to various matting encoders, its initial predictions outperform the corresponding baselines. This result is surprising, as the model uses only white noise as guidance in the first step, similar to one-step methods. We attribute this improvement to our UTI training strategy. By forcing the model to overcome the errors in the guidance, the model undergoes more thorough training.

\begin{figure}
    \centering
    \includegraphics[width=0.95\textwidth]{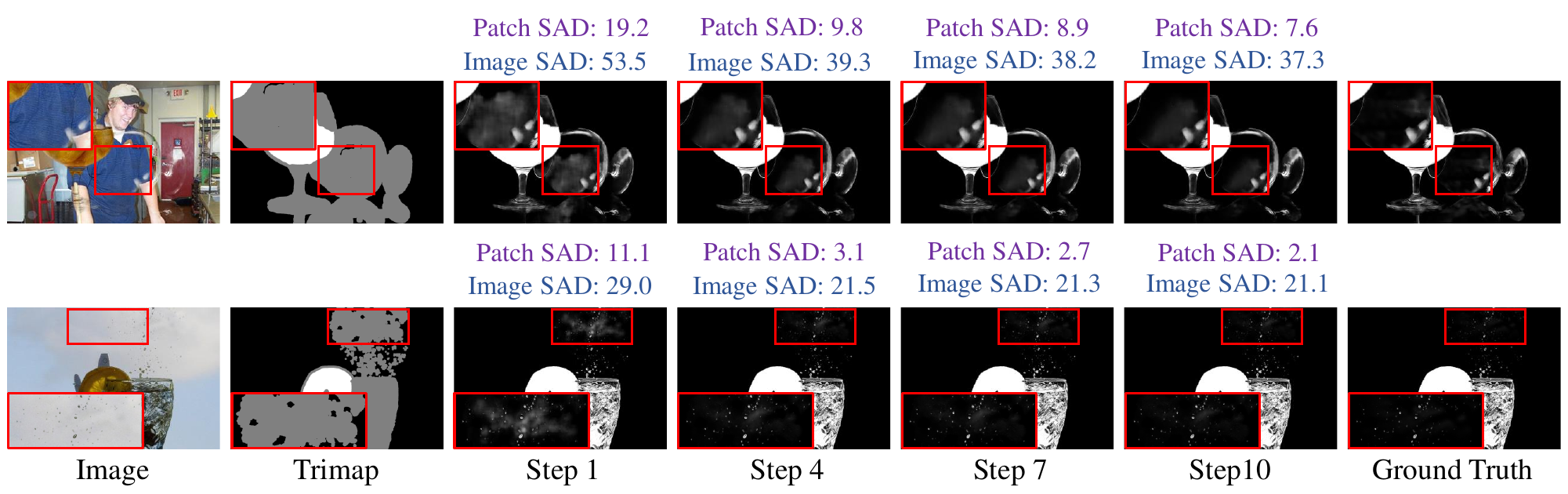}
    \caption{Visualization of the iterative refinement. DiffMatte tends to correct error-intensive local areas. Patch indicates the area within the red box. }
    \label{fig:refine}
\end{figure}

\noindent\textbf{Performance gain with iteration.} The results in \ref{sec:result} demonstrate that DiffMatte can progressively enhance the alpha matte. As shown in Fig.\ref{fig:refine}, by examining the results of each iteration, we believe DiffMatte achieves this by focusing on \textbf{error-intensive} areas in the predictions. Although noise samples do not explicitly guide the model to optimize specific areas, DiffMatte's iterative process allows the model to autonomously correct errors. Consequently, DiffMatte helps improve performance on weaker encoders like ResNet-34 and Swin-Tiny. For stronger encoders in the ViT series, the performance gains from local error correction are averaged across the test set, thus less noticeable in the curve. \\

\subsection{Ablation Study}
\label{sec:ablation}

We first perform ablation experiments on our denoising diffusion approach on the Composition-1k test set. All models are trained using ViT-S as the encoder. 

\noindent\textbf{Effect of Diffusion Process.} We ablate the noises in our diffusion framework to explore its effects. We first remove the noise term of DiffMatte, \textit{i.e.}, $\epsilon$ in the Eq.\ref{eq:2}, and predict alpha matte starting from an empty image during inference. This practice makes the model overly dependent on the direct guidance of clean alpha matte, causing catastrophic inference failures that cannot be fixed even with self-aligned training. We also try to fix this issue with iterative training on the existing one-step matting method. We modified ViTMatte to accept an additional 1-channel alpha matte and constrained two consecutive predictions during retraining, using an empty image and the first prediction for guidance. The results in the second row of Tab.\ref{tab:nands} show that iterative ViTMatte is still affected by errors in previous predictions, leading to performance degradation. Additionally, the first step results are weaker than the non-iterative manner, likely due to the large gap in guidance information. In summary, noise in iterative matting disrupts previous predictions and prevents over-reliance, allowing the model to correct errors through denoising and improving matting quality.

\setlength{\intextsep}{-4pt}
\begin{wrapfigure}{r}{0.42\textwidth}
    \centering
    \includegraphics[width=0.4\textwidth]{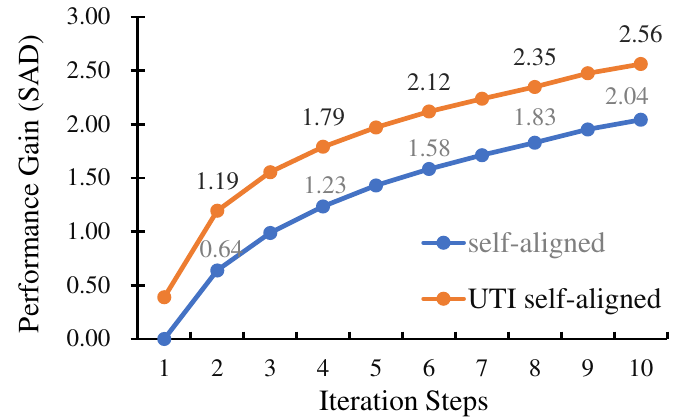}
    \caption{Performance gain compared with no self-aligned adopted.}
    \label{fig:self-aligned}
\end{wrapfigure}
\begin{table}[t!]
    \centering
    \setlength\tabcolsep{6pt}
    \renewcommand\arraystretch{1.4} 
    \caption{Ablation on the role of noises and self-aligned strategy. $\textbf{iter}$ denotes the iterative modification. $i$ in $SAD_{i}/MSE_{i}$ represents the number of iteration steps. The original Self-aligned shows performance degradation after 10 steps.}
    \label{tab:nands}
    \scalebox{0.8}{
    \begin{tabularx}{\textwidth}{l|cccc}
        \toprule
         Method & SAD$_1$/MSE$_1$ & SAD$_5$/MSE$_5$ & SAD$_{10}$/MSE$_{10}$ & SAD$_{15}$/MSE$_{15}$ \\
        \midrule
         ViTMatte~\cite{yao2023vitmatte} & 21.46/3.30 & -/- & -/- & -/- \\
         ViTMatte $\textbf{iter}$ & 23.87/4.21 & 24.19/4.38 & 24.84/4.79 & 25.12/5.08 \\
        \midrule
        \underline{\textit{Diffusion}} &  &  &  &  \\
        w\textbackslash o  Self-aligned & 21.05/3.16 & 22.51/3.37 & 23.04/4.07 & 23.68/4.21 \\
        Self-aligned~\cite{ji2023ddp} & 21.11/3.21 & 21.08/3.17 & 21.04/3.19 & 21.06/3.22 \\
        \midrule
        UTI Self-aligned & \textbf{20.61}/\textbf{3.08} & \textbf{20.54}/\textbf{3.07} & \textbf{20.52}/\textbf{3.06} & \textbf{20.50}/\textbf{3.04} \\
        \bottomrule
    \end{tabularx}}
\end{table}
\noindent\textbf{Effect of Self-aligned Strategies.} As shown in Tab.\ref{tab:nands}, the matting performance will gradually decrease without using any self-aligned strategy. This phenomenon has been explained in Sec.\ref{sec:iterative}. When using the self-aligned strategy, our UTI has higher single-step performance and more sustained iterative improvement compared with the approach proposed in \cite{ji2023ddp}. This is because the noise samples provided by UTI are more abundant, allowing the model to cope with different levels of errors, and thus better adapt the guidance provided by the noise samples. Fig.\ref{fig:self-aligned} intuitively shows the improvement brought by the self-aligned strategy and the superiority of UTI.

\noindent\textbf{Effect of Noisy Schedule.} We study the effect of the noisy schedule in Tab.\ref{tab:noise} and observe that the linear schedule is best suited for matting tasks compared to cosine and sigmoid schedules. This conclusion differs from the practice of diffusion methods on generative \cite{chen2023importance} and other vision tasks \cite{ji2023ddp}. This may attributed to the need for fine area guidance, as the linear schedule's relatively stable SNR variation ratio facilitates the model for detail perception. 

\noindent\textbf{Effect of Input Scaling.} We conduct the ablations on the input scaling and the results are shown in Tab.\ref{tab:scale}. As the factor decreases from $1$ to $0.01$, the performance of DiffMatte shows a trend of first increasing and then decreasing, finally reaching the optimum at $0.2$. We believe that appropriately reducing the scaling factor can increase the difficulty of the model extracting information from noisy samples and help the model learning. However, a large scaling factor destroys too much information and reduces performance.

\noindent\textbf{Effect of Diffusion Decoder Channels.} As shown in Tab.\ref{tab:channel}, we study the role of the number of decoder channels $N_d$. The hyperparameter $N_d$, which controls the number of $C_{out}$ channels of the convolutional layer in Up-Block and Down-Block pairs, can be varied to meet accuracy or recourses requirements. As $N_d$ increases, the matting accuracy of the model will be improved, but the number of parameters and computational overhead will also increase. We find that when $N_d$ is set to 32, it can best balance the performance and costs.

\begin{table}[t]
\centering
\hspace{-0.5em}
\caption{Ablation experiments with ViT-S~\cite{yao2023vitmatte} on Composition-1k test set~\cite{xu2017deep}. We report the performance with 10 steps. Default settings are marked in gray.}
\subfloat[
    \scriptsize \textbf{Noise Schedule}. We find linear schedule works best for matting task.
    \label{tab:noise}
]{
    \begin{minipage}{0.28\linewidth}{
        \begin{center}
            \begin{tabular}{l|cc}
            \toprule
            \scriptsize
            Noise Schedule & SAD & MSE  \\
            \hline
            sigmoid & 21.46 & 3.31 \\
            cosine & 21.58 & 3.35 \\
            \rowcolor{gray!20} 
            \textbf{linear} &  \textbf{20.52} & \textbf{3.06}  \\
            \bottomrule
            \multicolumn{3}{c}{~}\\
            \multicolumn{3}{c}{~}\\
            \end{tabular}
        \end{center}}
    \end{minipage}
}
\hspace{1em}
\subfloat[
    \scriptsize \textbf{Input Scaling}. The best input scaling factor for DiffMatte is 0.2.
    \label{tab:scale}
]{
    \centering
    \begin{minipage}{0.26\linewidth}{
        \begin{center}
            \begin{tabular}{c|cc}
                \toprule
                \scriptsize
                Input Scaling & SAD & MSE \\
                \hline
                0.05 & 21.0 & 3.32 \\
                0.1 & 21.30 & 3.24 \\
                \rowcolor{gray!20}
                \textbf{0.2} & \textbf{20.52} & \textbf{3.06} \\
                0.5 & 20.88 & 3.24 \\
                1.0 & 22.19 & 3.75 \\
                \bottomrule
            \end{tabular}
        \end{center}}
    \end{minipage}
}
\hspace{1em}
\subfloat[
    \scriptsize \textbf{Diffusion Decoder Channels}. $N_d$ set to 32 can best balance the parameters and performance.
    \label{tab:channel}
]{
    \begin{minipage}{0.33\linewidth}{
        \begin{center}
            \begin{tabular}{c|cccc}
                \toprule
                \scriptsize
                $N_d$ & SAD & MSE & Params & TFLOPs \\
                \hline
                16 & 21.63 & 3.55 & 1.9M & 0.2 \\
                24 & 21.23 & 3.27 & 3.4M & 0.5 \\
                \rowcolor{gray!20}
                32 & 20.52 & 3.06 & 5.3M & 0.8 \\
                48 & 20.29 & 3.05 & 10.0M & 1.8\\
                \bottomrule
                \multicolumn{4}{c}{~}\\
            \end{tabular}
        \end{center}}
    \end{minipage}
}
\hspace{1em}


\label{tab:ablations}
\end{table}
	\section{Conclusion}
\label{sec:conclusion}
In this paper, we proposes an iterative natural image matting framework by introducing the denoising diffusion process. Our DiffMatte achieves new state-of-the-art performances on the Composition-1k, and beats the corresponding baseline on the generalization test and AIM-500 benchmark. By iteratively revising alpha matte, DiffMatte can improve prediction quality with low computational overhead and running time. This contributes to our decoupled decoder design and UTI self-aligned training strategy. We hope that DiffMatte can promote research on interactive matting and lead to practical image editing applications.

        \section*{Acknowledgements}
        This research was funded by the Fundamental Research Funds for the Central Universities (No. 2023JBZD003, No. 2022XKRC015), and supported by the National NSF of China (No.U23A20314).

	%
	%
	\bibliographystyle{splncs04}
	\bibliography{main}

	\newpage
	\begin{center}
	\vspace{1em}
		\Large
		\textbf{Appendix}
	\vspace{1em}
	\end{center}
	\appendix

\section{Discussions}
\subsection{Diffusion for Perception.}
Denoising diffusion theory \cite{ho2020denoising,song2020denoising,song2020score} are proven to be dominantly effective in image generation and related tasks \cite{rombach2022high,zhang2023adding}. However, its role in image perception tasks has not been fully explored. When applying the diffusion process to the matting task, we aim to reuse the alpha matte to improve the prediction quality in the next step. Under this setting, the noise sample $X_t$ at each step $t$ is formed by the previous steps' predicted alpha matte $X_0$, and the prediction target is the clean alpha matte instead of the current step noise $\epsilon$. This manner is proven to be effective in various diffusion-for-perception methods~\cite{amit2021segdiff,chen2023diffusiondet,chen2023generalist,ji2023ddp}, which is contrary to the practice used for image generation. The reason lies that the perception task requires image-specific features, and predicting noise is not conducive to training the feature extraction of the model.
\begin{wrapfigure}[21]{r}{0.48\textwidth}
    \centering
    \includegraphics[width=0.48\textwidth]{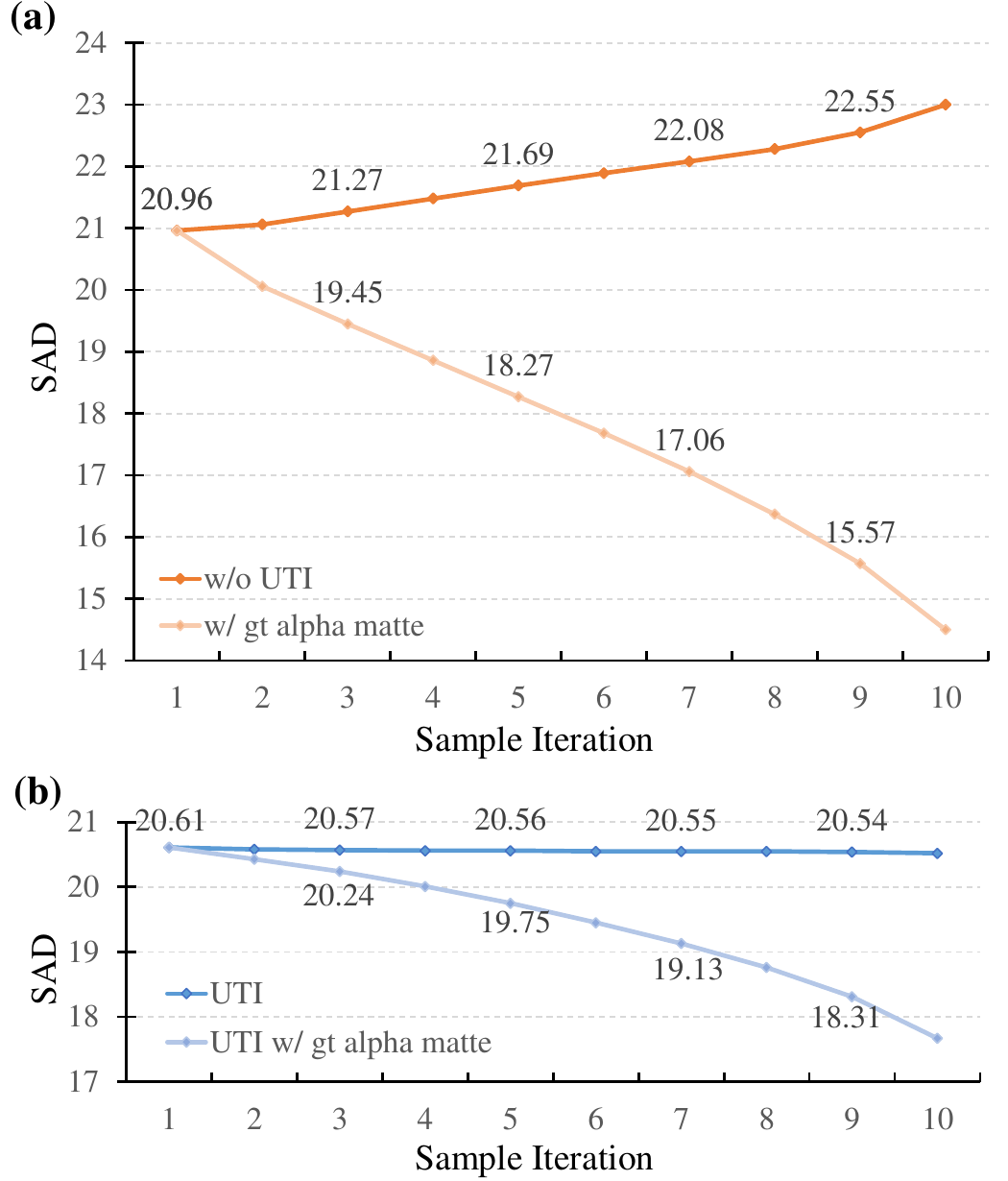}
    \caption{\textbf{Consistent Sampling.} Using consistent sampling versus inconsistent sampling. (a) Model without training with UTI. (b) Model training with UTI.}
    \label{fig:c1k}
\end{wrapfigure} 
In our practice, the denoising process from Gaussian noise to clear alpha matte is completed in several DDIM steps. DDIM steps is a specific name for time steps when using the DDIM sampling strategy. Since we use a continuous-time training method, we can arbitrarily specify DDIM steps during inference without retraining the model. After the total number of DDIM steps $T$ is given during inference, we specify each reverse process in the diffusion process of $T$ steps as a single sampling step. 

\subsection{Consistent Sampling.} 
We discussed the negative impact of inconsistent sampling on the diffusion process in the main paper, and give a more intuitive explanation here. We try to use ground truth alpha matte to add noise in each sampling iteration to obtain noise samples during inference to eliminate the impact of previous prediction errors on subsequent predictions and align the sampling during training and inference. In Figure A (a) we use the model without UTI self-align, and in Figure A (b) we use the model trained with UTI. We found that using consistent sampling can more obviously improve the accuracy of matting as the number of iterations increases, but because this inconsistency cannot be eradicated, the matting performance of models without UTI decreases. Models using UTI can alleviate the trend of performance degradation to a certain extent, but there is still a big gap in the performance of consistent sampling. At the same time, we found that the model trained using UTI performed worse than the other one with consistent sampling. We believe that UTI training provides the model with resistance to noisy samples, so that it does not rely entirely on previous predictions to get the results of the current iteration.

\section{Decoder Structure Details}

\begin{table}[t]
    \centering
    \scriptsize
    \setlength{\tabcolsep}{0.5mm}{
    \caption{Hypermeters for Adobe Image Matting training. The scheduler value indicates the set of learning rates in different training stages. The scheduler milestone is the set of transition timings (epoch) between stages. $^*$ means using images of size $1024 \times 1024$ for training.}
    \label{tab:hypermeters}
    \vspace{-5pt}
    \begin{tabular}{l|cccccc}
        \toprule
        Matting Encoder & initial $lr$ & epochs & batch size & scheduler value & scheduler milestone & GPUs(A6000) \\
        \hline
        ResNet-34~\cite{li2020natural} &  $4e-4$ & 150 & 20 & [0.1, 0.05, 0.01] & [30, 90, 140] & 2 \\
        Swin-Transformer~\cite{park2022matteformer} &  $4e-4$ & 150 & 20 & [0.1, 0.05, 0.01] & [30, 90, 140] & 2 \\
        ViT-S~\cite{yao2023vitmatte} & $5e-5$ & 150 & 16 & [0.1, 0.05] & [30, 90] & 2\\
        ViT-S$^*$~\cite{yao2023vitmatte} & $5e-5$ & 200 & 8 & [0.1, 0.05, 0.01] & [60, 110, 170] & 4\\
        ViT-B~\cite{yao2023vitmatte} &  $2.5e-4$ & 170 & 16 & [0.1, 0.05, 0.01] & [30, 90, 140] & 2\\
        \bottomrule       
    \end{tabular}}
    \vspace{-3pt}
\end{table}

\begin{table}[t]
    \centering
    \scriptsize
    \setlength{\tabcolsep}{0.5mm}{
    \caption{Hypermeters for Distinctions-646 training. The scheduler value indicates the set of learning rates in different training stages. The scheduler milestone is the set of transition timings (epoch) between stages.}
    \label{tab:hypermeters-d646}
    \vspace{-5pt}
    \begin{tabular}{l|cccccc}
        \toprule
        Matting Encoder & initial $lr$ & epochs & batch size & scheduler value & scheduler milestone & GPUs(A6000) \\
        \hline
        ViT-S~\cite{yao2023vitmatte} & $5e-5$ & 150 & 16 & [0.1, 0.05] & [30, 90] & 2\\
        ViT-B~\cite{yao2023vitmatte} &  $2.5e-4$ & 200 & 16 & [0.1, 0.05, 0.01] & [30, 90, 140] & 2\\
        \bottomrule 
    \end{tabular}}
    \vspace{-10pt}
\end{table}
The scale of the top-level features of the matting encoder may be 1/32 (normal hierarchical backbone network) or 1/16 of the input image size (non-hierarchical backbone network or using the Atrous Spatial Pyramid Pooling module). In order to adapt to various matting encoders, we design customized generation rules for the decoder. Given the input channel $N_c$ and $N_{F}$ of $c$ and $F$, we introduce a hyperparameter $N_{d}$, which is a relatively small number, to control the channel numbers of Down-block and Up-block. $\{ N_c, N_{d}, 2 \times N_{d}, 4 \times N_{d}, \textcolor{blue}{4 \times N_{d}} \}$ and $\{ N_{F}, \textcolor{blue}{8 \times N_{d}}, 8 \times N_{d}, 4 \times N_{d}, 2 \times N_{d}, N_{d} \}$ are the channel number lists of Down-Block and Up-Block respectively, and an \textcolor{blue} {additional pair of blocks} are stacked when dealing with the feature size of 1/32 of the input image. The exchange of texture information between Down-block and Up-block is accomplished by simple feature concatenation. A matting head is added after the last Up-Block to obtain a one-channel alpha matte. The two kinds of blocks additionally incorporate the encoding and fusion of time step information to make our diffusion decoder time-aware. This is achieved by a linear layer that learnably transforms $t$ values into high-dimensional temporal embeddings $\tau$. $\mathcal{D}$ is constructed only by CNN layers since its role is to refine the texture information during the diffusion process and therefore does not need to use any attention mechanism.

\section{Implementation Details}
\noindent\textbf{Adobe Image Matting.} Following the common practice \cite{xu2017deep, forte2020f, park2022matteformer}, the training image is cropped to $512 \times 512$ after synthesizing by random foreground and background, and the corresponding trimap is generated through a dilation-erosion operation using a random kernel size of $[1,30]$. We implement our DiffMatte based on several encoders redesigned for matting tasks, including Resnet-34 \cite{li2020natural}, Swin-Transformer \cite{park2022matteformer}, ViT-S and ViT-B \cite{yao2023vitmatte}. We use the AdamW optimizer for training and incorporate the weight decay trick as well as a multi-stage learning rate adjustment strategy. Notably, to make a fair comparison with \cite{liu2023rethinking}, we additionally use input images of size $1024 \times 1024$ for training. Hyperparameters in the training phase vary for models with different settings, as be explained in Tab.\ref{tab:hypermeters}. 

\noindent\textbf{Distinctions-646.} The D646 dataset provides 646 unique foregrounds and corresponding alpha matte. It is divided into a training set and a test set, which contain 596 training foregrounds and 50 test foregrounds respectively. Data augmentation during training is consistent with the operation on the Adobe Image Matting dataset. We trained DiffMatte-ViTS and DiffMatte-ViTB on D646, and the training parameter settings are shown in Tab.\ref{tab:hypermeters-d646}.

\section{Experimental Results}

\begin{table}[t]
    \centering
    \setlength{\tabcolsep}{1.mm}{
    \caption{Quantitative results on Distinctions-646 ~\cite{qiao2020attention}. S1 and S10 denote 1 and 10 DDIM time steps.}
    \label{tab:supp-d646}
    \vspace{-2pt}
    \begin{tabular}{l|cccc}
        \toprule
        Methods & SAD & MSE($10^3$) & Grad & Conn \\
        \hline
        ViTMatte-S~\cite{yao2023vitmatte} & 20.57 & 6.12 & 10.06 & 16.97 \\
         \rowcolor{gray!15} DiffMatte-ViTS (S1) & 18.93 & 5.90 & 10.17 & 16.28 \\
         \rowcolor{gray!15} DiffMatte-ViTS (S10) & \textbf{18.91} & \textbf{5.81} & \textbf{10.00} & \textbf{16.26} \\
         \hline
        ViTMatte-B~\cite{yao2023vitmatte} & 16.22 & 4.51 & 7.53 & 13.56 \\
         \rowcolor{gray!15} DiffMatte-ViTS (S1) & 15.47 & 4.24 & 7.18 & \textbf{13.20} \\
         \rowcolor{gray!15} DiffMatte-ViTS (S10) & \textbf{15.43} & \textbf{4.21} & \textbf{6.95} & 13.22 \\
        \bottomrule
    \end{tabular}}
    \vspace{-2pt}
\end{table}
\begin{table}[t]
    \centering
    \setlength{\tabcolsep}{1.mm}{
    \caption{Quantitative results on Composition-1K ~\cite{xu2017deep}. S1 and S10 denote 1 and 10 DDIM time steps.}
    \label{tab:supp-c1k}
    \vspace{-3pt}
    \begin{tabular}{l|cccc}
        \toprule
        Methods & SAD & MSE($10^3$) & Grad & Conn \\
        \hline
        AEMatter~\cite{liu2023rethinking} & 17.79 & 2.39 & \textbf{4.81} & 12.64 \\
         \rowcolor{gray!15} DiffMatte-ViTS (S1) & 17.16 & 2.26 & 5.14 & 11.42 \\
         \rowcolor{gray!15} DiffMatte-ViTS (S10) & \textbf{17.15} & \textbf{2.26} & 5.13 & \textbf{11.42} \\
        \bottomrule
    \end{tabular}}
    \vspace{-2pt}
\end{table}

In Tab.\ref{tab:supp-d646}, we provide results on the test set of D646. We compare the performance of DiffMatte-ViTS and DiffMatte-ViTB with ViTMatte-S and ViTMatte-B on official weights trained on the D646 training set. It can be observed that DiffMatte still outperforms ViTMate under the same backbone. Besides, after training on the D646 training set, DiffMatte adapts to the data domain and can achieve performance improvement as the number of DDIM steps increases. In Tab.\ref{tab:supp-c1k}, we present results on large-size training inputs. Our method surpasses AEMatter~\cite{liu2023rethinking} in most metrics, further demonstrating the effectiveness of our method.

\section{More Visualization Results}
\begin{figure}
    \vspace{0.5em}
    \centering
    \includegraphics[width=0.98\textwidth]{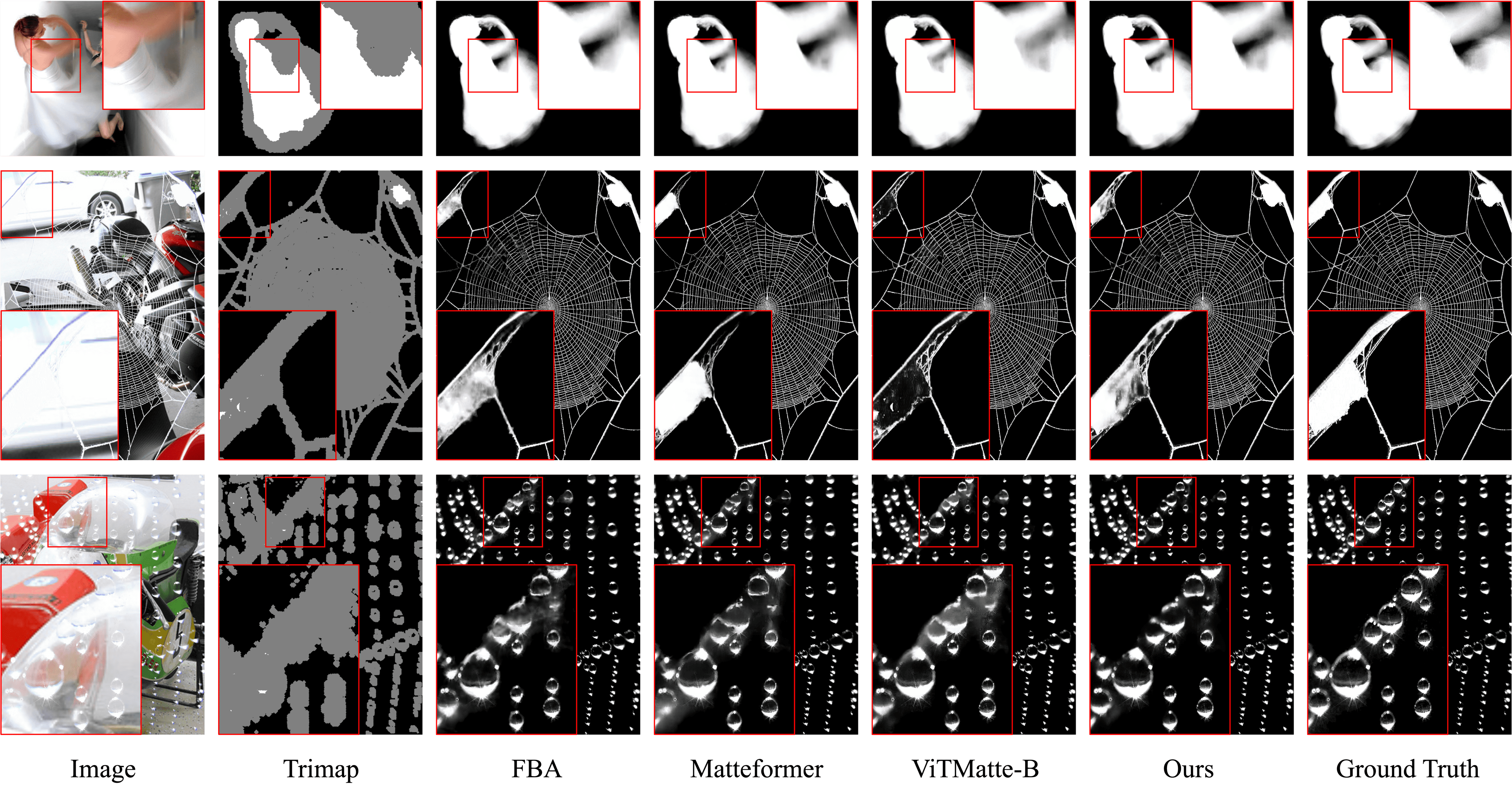}
    \vspace{-0.8em}
    \caption{Qualitative generalization results compared with previous SOTA methods on Distinctions-646 test set\cite{qiao2020attention}. Best viewed by zooming in.}
    \label{fig:d646}
    \vspace{-1.5em}
\end{figure}

\begin{figure}
    \vspace{0.5em}
    \centering
    \includegraphics[width=0.98\textwidth]{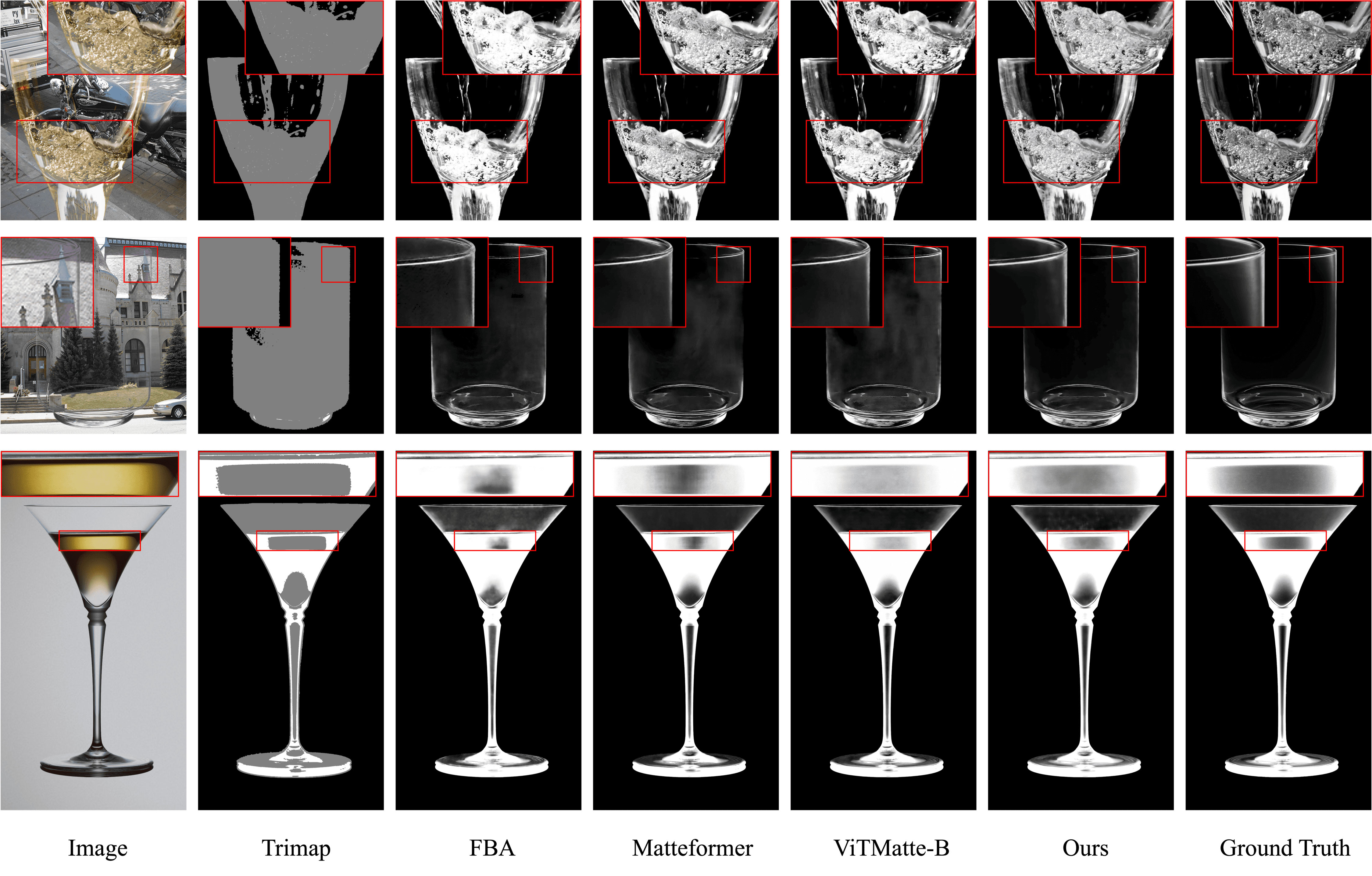}
    \vspace{-0.8em}
    \caption{Qualitative generalization results compared with previous SOTA methods on Semantic Image Matting test set \cite{sun2021semantic}. Best viewed by zooming in.}
    \label{fig:simd}
    \vspace{-1.5em}
\end{figure}

Fig.\ref{fig:d646} and \ref{fig:simd} show the comparison of DiffMatte (ViT-B) generalized to D646 and SIMD at 10 DDIM steps with previous state-of-the-art methods. These results show the performance advantages of our approach.

\end{document}